\documentclass[10pt,twocolumn,letterpaper]{article}

\usepackage{cvpr}              

\usepackage{arydshln}
\usepackage{pifont}
\usepackage[dvipsnames]{xcolor}
\usepackage{subcaption}
\usepackage{lipsum}
\usepackage{float}
\usepackage{graphicx}
\usepackage{marvosym}

\definecolor{cvprblue}{rgb}{0.21,0.49,0.74}
\usepackage[pagebackref,breaklinks,colorlinks,allcolors=cvprblue]{hyperref}
\def\paperID{22561} 
\def\confName{CVPR}
\def\confYear{2026}

\title{Exploring the best way for UAV visual localization under Low-altitude Multi-view Observation Condition: a Benchmark}

\author{
Yibin Ye\textsuperscript{1}\thanks{Co-first author.}\quad
Xichao Teng\textsuperscript{1}\footnotemark[1]\quad
Shuo Chen\textsuperscript{1}\quad
Leqi Liu\textsuperscript{1}\quad
Kun Wang\textsuperscript{1}\quad
Xiaokai Song\textsuperscript{1}\quad
Zhang Li\textsuperscript{1,\Letter}\\
\textsuperscript{1}National University of Defense Technology, China\\
{\tt\small zhangli\_nudt@163.com}
}

\begin{document}
\maketitle
\begin{abstract}
Absolute Visual Localization (AVL) enables an Unmanned Aerial Vehicle (UAV) to  determine  its position in GNSS-denied environments  by establishing geometric relationships between UAV images and  geo-tagged reference maps. While many previous works have achieved AVL with image retrieval and matching techniques, research in low-altitude multi-view scenarios still remains limited. \textbf{Low-altitude} \textbf{Multi-view} conditions present greater challenges due to extreme viewpoint changes. To investigate effective UAV AVL approaches under such conditions, we present this benchmark. Firstly, a large-scale low-altitude multi-view dataset called \textbf{AnyVisLoc} was constructed. This dataset includes 18,000 images captured at multiple scenes and altitudes, along with 2.5D reference maps containing aerial photogrammetry maps and historical satellite maps. Secondly, a unified framework was proposed to integrate the state-of-the-art AVL approaches and comprehensively test their performance. The best combined method was chosen as the baseline and the key factors influencing localization accuracy are thoroughly analyzed based on it.  This baseline achieved a 74.1\% localization accuracy within 5m under low-altitude, multi-view conditions. In addition, a novel retrieval metric called PDM@K was introduced to better align with the characteristics of the UAV AVL task. Overall, this benchmark revealed the challenges of low-altitude, multi-view UAV AVL and provided valuable guidance for future research. The dataset and code are available at \url{https://github.com/UAV-AVL/Benchmark}
\end{abstract}    
\section{Introduction}
\label{sec:intro}
\begin{figure}[!t]
    \centering
    \includegraphics[width=0.4\textwidth]{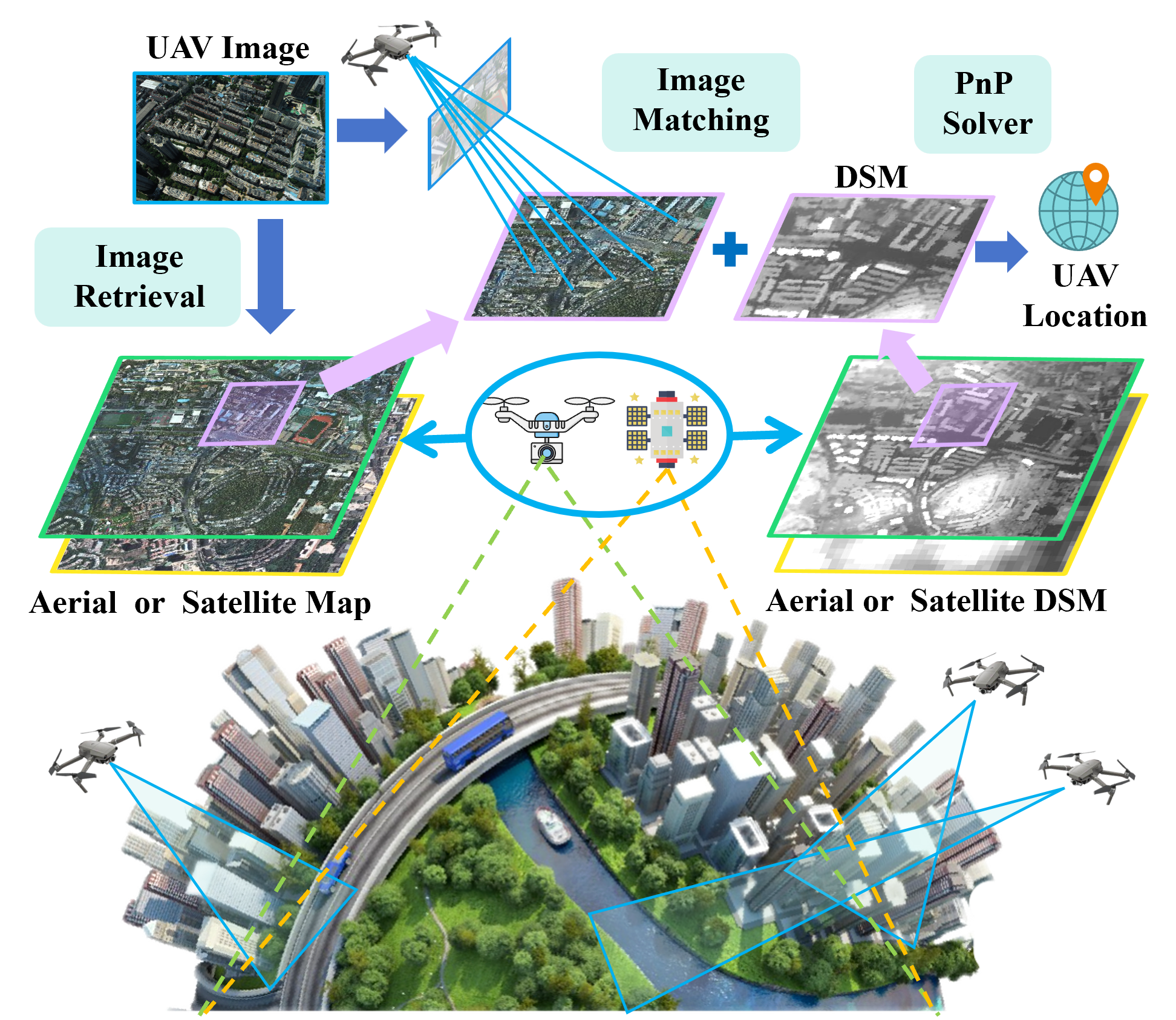}
    \caption{\textbf{Benchmark Overview. }This benchmark focuses on UAV visual localization under low-altitude multi-view conditions using the 2.5D aerial or satellite reference maps. The visual localization is mainly achieved via a unified framework combining image retrieval, image matching, and PnP problem solving.}
    \label{fig:abstract}
\end{figure}

In recent years, Unmanned Aerial Vehicles (UAVs) have been widely used in various fields such as precision agriculture, emergency response, and transportation \cite{couturier2021review}. These drones generally rely on  Global Navigation Satellite System (GNSS) and inertial navigation system (INS) for localization. However, GNSS signals are prone to dropout and jamming \cite{GPS_jamming}, while INS suffers from drifting over time \cite{he2024leveraging}. Consequently, there is a growing interest in visual localization as a temporary alternative \cite{he2023foundloc, DenseUAV,  he2024leveraging, chang2023review}. 

Current UAV visual localization research can be primarily categorized into relative visual localization (RVL) and absolute visual localization (AVL) \cite{chen2023oblique}. RVL approaches, such as Simultaneous Localization and Mapping (SLAM) \cite{ORB-SLAM3}, are commonly used for short-range robot navigation through frame-to-frame matching \cite{chang2023review}. While RVL methods do not require geo-tagged maps, they also suffer from drift\footnote{Note that Loop closure can somewhat improve global localization accuracy, but it is hard to achieve for long-distance outdoor flights \cite{couturier2021review}.}. In contrast, AVL approaches, which involve matching UAV images with geo-tagged reference maps, have inherent immunity to drift and are the focus of this paper. 
 
To the best of our knowledge, most previous studies focused on nadir-view-based AVL \cite{ bianchi2021uav, mughal2021assisting, schleiss2022vpair, he2023foundloc, GE_simulator}. Besides, the regions observed for AVL are often assumed as planar, in which the elevations are significantly lower than the UAV flight altitude \cite{UAV-VisLoc}. Therefore, one can use affine model to estimate the transformation between UAV images and geo-tagged reference maps \cite{song2019oblique}. However, as the significant viewpoint changes at low altitude introduce severe nonlinear transformation between UAV images and reference maps, this assumption does not apply to UAV  visual localization under low-altitude ($<300m$ \footnote{According to relevant regulations, civilian UAVs are not permitted to fly above 300 meters \cite{interim_regulations_uav}}) and multi-view (with pitch angles \footnote{the angle between the optical axis and the horizontal plane} ranging from 20° to 90°) conditions. These conditions present greater challenges for visual localization \cite{ye2024coarse, yin2023isimloc} and how current AVL methods perform in these challenging scenarios is still unclear. Meanwhile, no publicly available datasets could be used to comprehensively benchmark current AVL methods. To fill this gap, we  constructed a new dataset called AnyVisLoc and conducted extensive experiments on this dataset to evaluate all UAV AVL methods under low-altitude multi-view conditions. Overall, our contributions are as follows:
\begin{itemize}
	\item The first large-scale low-altitude multi-view dataset called AnyVisLoc was proposed. This dataset contains 18,000 UAV images captured at multiple scenes and altitudes, along with two types of 2.5D reference maps including aerial photogrammetry maps and satellite maps.
	\item A unified framework was introduced to integrate the state-of-the-art AVL approaches and thoroughly test their performance. The best combined method was chosen as the baseline and the key factors influencing baseline accuracy are thoroughly discussed, including the type of reference maps, different pitch angles and the noise in prior information.
	\item  A novel retrieval metric called PDM@K was designed to bridge the relationship between image retrieval accuracy and final localization accuracy. Compared to other metrics like Recall@K and SDM@K, PDM@K aligns more closely with the characteristics of the UAV AVL task.

\end{itemize}

\section{Related Work}
\label{sec:formatting}
\begin{table*}[h]
  \caption{\textbf{UAV AVL Datasets Comparison}}
  \label{tab:datasets}
  \centering
  \scalebox{0.69}{
  \begin{tabular}{lcccccccccc}   
    \toprule
    \textbf{Name} & \textbf{Year} & \textbf{UAV data} & \textbf{UAV Types} & \textbf{UAV Location} & \textbf{2D Reference}  & \textbf{DSM}  & \textbf{Flight Altitude} & \textbf{Observation} & \textbf{Scenes}	& \textbf{Temporal}
\\
    \midrule
    
    ATM \cite{mughal2021assisting}& 2021 & Real-scenes & 1 & \checkmark& Aerial  & $\times$ & unknown & Nadir-view & Urban & Multiple\\

    University-1652 \cite{University-1652} & 2021 & Synthetic & - & $\times$& Satellite & $\times$ & 121.5m to 256m & Multi-view & Buildings & -\\ 

     Wildnav \cite{gurgu2022vision} & 2022 & Synthetic & - & \checkmark& Satellite  & $\times$ & unknown & Nadir-view & Wilderness & - \\
     VPAIR \cite{schleiss2022vpair} & 2022 & Real-scenes  &1 & \checkmark & Satellite & $\checkmark $ & 300m to 400m & Nadir-view & Multiple & -\\
         
    SEUS-200 \cite{sues200} & 2023 & Real-scenes & -& $\times$ & Satellite  & $\times$ & 150/200/250/300m & Multi-view & Urban & -\\ 
       

    DenseUAV \cite{DenseUAV} & 2024 & Real-scenes  &1 & \checkmark & Satellite  & $\times$ & 80m/90m/100m & Nadir-view & Urban & Multiple\\
    UAVD4L \cite{Wu2023uav4l} & 2024 & Real-scenes & 1 & \checkmark & Aerial  & \checkmark & 50m to 151m & Multi-view & Urban & -\\
    UAV-VisLoc \cite{UAV-VisLoc} & 2024 & Real-scenes & unknown & \checkmark & Satellite  & $\times$ & 400m to 2000m & Nadir-view & Multiple & Multiple\\
    GTA-UAV \cite{ji2024game4loc} & 2025 & Synthetic & - & \checkmark & Satellite & $\times$ &
    80m to 650m & Near nadir-view & Multiple & -\\
    CVGL-RGBT \cite{zhou2025cdm} & 2025 & Real-scenes & 1 & $\times$ & Satellite & $\times$ & unknown & Multi-view & Multiple & Multiple \\
    \textbf{AnyVisLoc (Ours)} & 2026 & \textbf{Real-scenes} & \textbf{7} & \textbf{\checkmark }& \textbf{Aerial \& Satellite}  & \textbf{\checkmark }& \textbf{30m to 300m} & \textbf{Multi-view} & \textbf{Multiple} & \textbf{Multiple} \\

  \bottomrule
  \end{tabular}}
\end{table*}

\textbf{Image-Level Retrieval} can be used  to estimate the rough position of UAV image on the reference map \cite{ chen2024fpi, ye2024coarse}. These approaches includes Template Matching (TM) \cite{NCC} and Visual Geo-localization (VG) \cite{berton2022deep} methods. All these methods rely on constructing feature patterns to measure similarity between drone (query) and reference (gallery) images.
The similarity metrics commonly used for TM (e.g., NCC \cite{NCC}, MI \cite{MI}) are not invariant to viewpoint differences,
making them less suitable for low-altitude, multi-view conditions. In contrast, deep learning-based VG methods are more adaptable to changes in viewpoint \cite{sues200}. NetVLAD \cite{netvlad} is a widely adopted VG baseline. In the field of drone-to-satellite retrieval, several studies concentrate on refining network architectures \cite{LPN,RK-Net,shen2023mccg}. For instance, MCCG \cite{shen2023mccg} incorporates a cross-dimensional interaction module and a multi-classifier structure to construct comprehensive feature representations. Additionally, some efforts are dedicated to enhancing training and sampling strategies \cite{FSRA,s4g,DAC,wu2024camp}. For example, Sample4Geo \cite{s4g} employs InfoNCE loss to increase negative samples for contrastive learning within one batch. Meanwhile, CAMP \cite{wu2024camp} introduces comparisons between different scenes from the same imaging platform within the same training batch, thereby increasing the number of negative samples for contrast.


\textbf{Pixel-Level Image Matching} first establishes feature point correspondences between UAV images and reference map, subsequently solving the PnP problem to determine the UAV's location \cite{gurgu2022vision, chen2023oblique}. 
Pixel-level image matching, which comprises sparse and dense approaches \cite{RCM}, always achieve higher precision than image-level retrieval.
The typical sparse approaches involve keypoint detection and keypoint matching steps. Traditional keypoint detectors like SIFT \cite{SIFT}, SURF \cite{surf}, and ORB \cite{ORB-SLAM3} are widely used in sparse matching but are sensitive to large viewpoint differences. The rise of deep learning has brought learning-based detectors and descriptors into the mainstream to cope with viewpoint changes \cite{D2-Net,detone18superpoint,alike,potje2024xfeat}. Another line of sparse approaches focuses on  learning-based global relationship modeling among keypoints, such as SuperGlue \cite{superglue}, Lightglue \cite{lightglue} and Omniglue \cite{omniglue}. Dense approaches \cite{sun2021loftr,DKM,roma,dedode} avoid keypoint detection and directly find dense correspondences between images, leading to significant improvements over sparse methods but with lower efficiency. Therefore, there is growing interest in creating lightweight and high-precision matching techniques \cite{potje2024xfeat,wang2024efficient, kim2024keypt2subpx,RCM}.

\textbf{UAV Visual Localization Dataset.} As summarized in \cref{tab:datasets}, most current UAV datasets focus on nadir-view and lack multi-view coverage \cite{ gurgu2022vision, DenseUAV, UAV-VisLoc, mughal2021assisting}. Some datasets are restricted to specific scenes (e.g., urban \cite{DenseUAV} or wilderness \cite{gurgu2022vision} areas). Some datasets use synthetic data (e.g., Google Map  Studio \cite{University-1652}) that do not fit the real application requirement. Although some recent geo-localization datasets include multi-view UAV images \cite{University-1652, sues200}, they often lack ground truth for UAV positions and continuous geographical coverage, which is not suitable for precise AVL.

Regarding reference maps, most datasets currently only have 2D satellite images and lack high-precision digital surface model (DSM) \cite{UAV-VisLoc}. Open-source products such as ALOS DSM 30m \cite{nikolakopoulos2020accuracy} and NASADEM 30m \cite{li2022global} have relatively low spatial resolutions and do not meet the  precise AVL requirements under low-altitude multi-view conditions.

 To the best of our knowledge, none of the existing datasets simultaneously provide aerial reference map and satellite reference map of the same area. The works most similar to ours are \cite{wang2018unmanned, song2019oblique,chen2023oblique,ye2024coarse}, although these literature explored the feasibility of AVL based on oblique-view, their test data lacked sufficient scene coverage, and the datasets and codes were not open-sourced yet.



\section{AnyVisLoc Dataset}

The accuracy of UAV AVL is affected by various factors including observation  conditions 
  (e.g., flight altitude, pitch angle, and field of view), flight environment (scene, weather, and illumination), and reference map characteristics (source, spatial resolution, and modality differences) \cite{ couturier2021review, ye2024coarse}. Our proposed AnyVisLoc dataset covers all possible factors for UAV AVL under low-altitude multi-view observation condition. This section gives a detailed description of it. 
\label{sec:Dataset}
\subsection{Data Acquisition and Processing}

We collected a total of 18,000 UAV images using seven different types of DJI drones\footnote{Mavic 2, Mavic 3, Mavic 3 Pro, Phantom 3, Phantom 4, Phantom 4 RTK, Mini 4 Pro}, whose cameras have different intrinsic parameters. As shown in \cref{fig:AVL_dataset}, the images were taken in 15 cities across China under different weather (sunny, cloudy, rainy), seasons, and illuminations (day and night), covering 25 distinct regions. The imaging coverage area ranges from 10,000 $m^2$ to 9,000,000 $m^2$.  
Each image is attached with camera intrinsic and extrinsic parameters, which can be used for UAV AVL as well as the ground truth. 

Besides the UAV images, two types of 2.5D reference maps are provided: \textbf{Aerial Photogrammetry Map} that is created by capturing images using the DJI drones and then applying modern Structure-from-Motion techniques \cite{Wu2023uav4l} to construct 2D orthomosaic and DSM. \textbf{Satellite Map} that involves 2D historical images from Google Earth and ALOS 30m DSM.
Both sets of 2.5D reference maps are geo-tagged and reprojected into the UTM coordinate system for geo-localization. 

Beyond these resources, we also provide high-precision correspondences between UAV images and the aerial photogrammetry map, enabling the community to train image retrieval and matching models specifically tailored for UAV AVL tasks. The correspondence establishment process and our training results are detailed in the appendix.
    \begin{figure}[!h]
    \centering
     \includegraphics[width=0.45\textwidth]{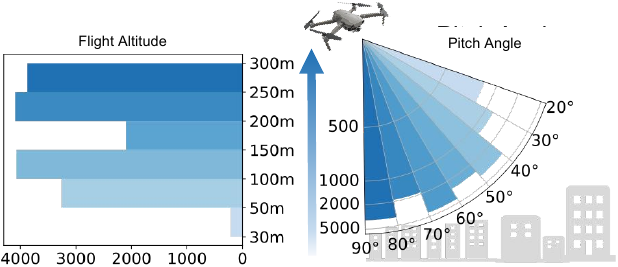}
    \caption{\textbf{Flight Altitude and Pitch Angle Distribution.} }
    \label{fig:pitch_and_alt}
\end{figure}
\subsection{Data Characteristics}
\begin{figure*}[!h]
    \centering
    \includegraphics[width=1\textwidth]{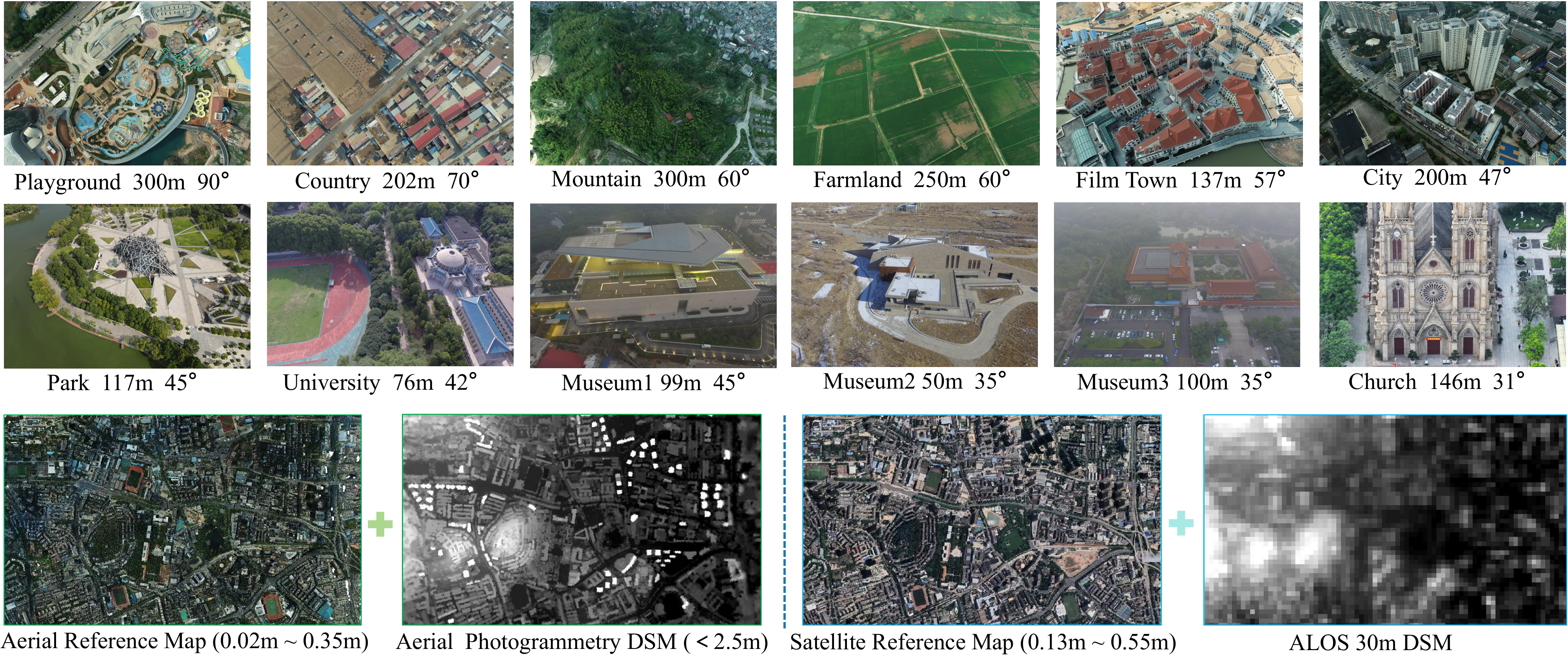}
    \caption{\textbf{Dataset Overview.}  The AnyVisLoc dataset contains multi-scene, multi-altitude, and multi-view UAV images  taken in 15 cities across China, as well as  aerial and satellite reference maps. Each UAV image shows its flight altitude and pitch angle below.}
    \label{fig:AVL_dataset}
\end{figure*}

\begin{itemize}
        \item \textbf{Multi-altitude}: Our dataset covers low-altitude flight conditions from 30m to 300m (see \cref{fig:pitch_and_alt} and \cref{fig:AVL_dataset}). 
        \item \textbf{Multi-view}:  Our dataset  covers pitch angle from 20° to 90° (see \cref{fig:pitch_and_alt} and \cref{fig:AVL_dataset}). 
        \item \textbf{Multi-scene}: Our dataset includes various scenes (see \cref{fig:AVL_dataset}), such as dense urban areas (e.g., cities and towns), typical landmark scenes (e.g., playground, museums, church), natural scenes (e.g., farmland and mountains), and mixed scenes (e.g., universities and  parks). 
        \item \textbf{Multi-reference map}: Our dataset provides two types of 2.5D reference maps for different purposes (see \cref{fig:AVL_dataset}). The aerial map with high spatial resolution can be used for high-precision localization but requires prior aerial photogrammetry. The satellite map serves as an alternative when the aerial map is unavailable. 
        
        \textbf{Overall}, our dataset can facilitate a comprehensive evaluation of current UAV AVL methods under low-altitude multi-view observation conditions. 
        
    \end{itemize}

\section{The Unified AVL Framework}
\label{sec:framework}
Given a UAV image captured under low-altitude, multi-view conditions, its field of view is within the reference map, but its exact position is unknown. Under such conditions, extreme viewpoint changes and a vast search space always result in image matching failures. Meanwhile, image retrieval can only provide a rough estimate of the current view’s position. Consequently, independently evaluating the performance of image retrieval or matching methods becomes challenging. Therefore, a unified framework is needed for fair and meaningful testing.

\textbf{Framework Workflow: }As shown in \cref{fig:abstract}, the framework operates in a coarse-to-fine manner. Initially, image-level retrieval is used to estimate the rough position of the current view, followed by pixel-level matching to obtain 2D-2D matched point pairs between the UAV image and the reference map. Subsequently, DSM data and the matching results are used to solve the PnP problem and determine the UAV’s geo-location. This framework can integrate with  different reference maps and localization strategies.

\textbf{Integrated Algorithms: }The image retrieval approaches include template matching approaches like NCC \cite{NCC} and MI \cite{MI}, popular retrieval models like NetVLAD \cite{netvlad}, and drone-to-satellite retrieval approaches such as CNN-based models \cite{LPN, shen2023mccg, s4g, RK-Net, MEAN, LRFR} and transformer-based models \cite{FSRA, DenseUAV, DAC, wu2024camp, QDFL}. Matching techniques include handcrafted methods such as SIFT \cite{SIFT} and ORB \cite{ORB-SLAM3}, learning-based sparse feature approaches \cite{D2-Net, alike, detone18superpoint, superglue, lightglue, omniglue, potje2024xfeat, kim2024keypt2subpx, LiftFeat}, (semi-) dense feature approaches \cite{sun2021loftr, tuzcuouglu2024xoftr, roma, DKM, dedode}, along with large-scale training frameworks \cite{xuelun2024gim, ren2025minima}. The PnP solver is P3P+RANSAC \cite{P3P}. 

Additionally, we compare four localization strategies: (a) Matching the Top-1 retrieval image \cite{ye2024coarse}, (b) Re-ranking Top-K retrievals based on the number of inliers 
 \cite{chen2021real}, (c) Matching all gallery images followed by most inliers sorting \cite{gurgu2022vision}, (d) Directly performs pixel matching between the UAV image and the reference map without retrieval \cite{yuan2020automated}.

\textbf{Localization Metric: }The localization accuracy is denoted as \scalebox{0.9}{$A@T=\frac{N_{T}}{N}\times 100\%$}, which means the correct localization ratio under a given threshold of $T$ meters.  The localization error is denoted as  \scalebox{0.85}{$e_i = \sqrt{(x_p - x_g)^2 + (y_p - y_g)^2}$}, where $(x_p,y_p)$ and $(x_g, y_g )$ are the UTM coordinates of predicted location and the ground truth, respectively. 

\textbf{Retrieval Metric: }The Recall@K and our proposed Pixel Distance Metric (PDM@K, see \cref{equ:PDM}) are used.  In \cref{equ:PDM}, $R_i$ denotes the overlap rate between retrievals and the ground truth, $R_i=d_i/(w_i\cdot r)$, where ${d_i}$ is the spatial distance, $w_i$ is the width of the gallery image, and $r$ is the spatial resolution of the reference map. $\frac{{{e^{ - \lambda  \cdot ({R_i} - \alpha )}}}}{{1 + {e^{ - \lambda  \cdot ({R_i} - \alpha )}}}}$ is the retrieval score of the i-th sample in the retrieval result order and $(K-i+1)$ is the weight. $\lambda $ and $\alpha$ are
weight factors. In this paper, $\lambda $ is set to 6 and $\alpha$ is set to 0.9. A larger value of PDM@K indicates a better retrieval performance. The rationale of this metric is detailed in \cref{sec:metric}.
\begin{equation}
\resizebox{.9\hsize}{!}{$PDM@K = \sum\limits_{i = 1}^K {\frac{{{{(K - i + 1)}} \cdot {e^{ - \lambda  \cdot ({R_i} - \alpha )}}}}{{1 + {e^{ - \lambda  \cdot ({R_i} - \alpha )}}}}} /{\sum\limits_{i = 1}^K {(K - i + 1)}}$.}
 \label{equ:PDM}
\end{equation}

\textbf{Implementation Details:} Our experiments were conducted on a machine equipped with Intel Core i9-10920X CPU @ 3.50GHz (128G RAM) and NVIDIA RTX 3090 GPU(24G). For deep learning-based image retrieval  approaches, we use the weight trained on the University-1652 dataset \cite{University-1652} with an image size of $384\times384$. For deep learning-based image matching approaches, we adopted weights and input sizes from existing works to leverage the models’ optimal performance while testing their generalization capabilities. For NCC and MI, we employed our own implementations. For SIFT, ORB and PnP solver, we used OpenCV implementations. Notably, we use the UAV’s altitude and pitch/yaw information to roughly estimate the image’s scale and rotation, which helps to narrow the search space. Further discussion on this setup is in \cref{sec:prior_inf}.

\section{Experiments on AVL Framework}
Based on the unified AVL framework, we evaluate state-of-the-art image retrieval and matching approaches alongside different localization strategies in \cref{sec:retrieval}, \cref{sec:matching}, and \cref{sec:strategy}, respectively. These experiments are conducted using a subset of UAV images and the aerial map. The best combined method is then selected as the baseline and evaluated on the complete dataset in \cref{sec:baseline}. We further present an analysis of our proposed PDM@K metric in \cref{sec:metric}.
\subsection{Image retrieval Approach}
\label{sec:retrieval}
\begin{table}[h]
\centering
\caption{\textbf{Performance Metrics of Retrieval Approaches}}
\scalebox{0.7}{
\begin{tabular}{l|c c c |c c c |c }
\hline\hline
Method & R@1 & R@3 & R@5 & P@1 & P@3 & P@5 & ms/feat \\
\hline\hline
NCC \cite{NCC} & 0.3 & 1.5 & 2.6 & 0.193     & 0.189     & 0.189 & 17$\dag$\\
MI \cite{MI} & 2.4 & 5.9 & 8.6 & 0.352     & 0.333     & 0.323 & 20$\dag$\\
\addlinespace[0.3em] 
 \hdashline
 \addlinespace[0.3em] 
NetVLAD \cite{netvlad} & 31.1 & 53.9 & 64.9 & 0.741     & 0.695     & 0.652 & 11\\
LPN \cite{LPN} & 41.1 & 62.1 & 70.4 & 0.799     & 0.719     & 0.665 & 5\\
RK-Net \cite{RK-Net} & 45.3 & 67.8 & 76.1 & 0.848     & 0.775     & 0.712  & 9\\

FSRA \cite{FSRA} & 52.9 & 73.9 & 81.2 & 0.870     & 0.780     & 0.717 & 5\\
DenseUAV \cite{DenseUAV} & 51.6 & 69.4 & 76.7 & 0.854     & 0.754     & 0.684 & 6\\

MEAN \cite{MEAN} & 48.0 & 70.8 & 77.1& 0.846     & 0.752     & 0.684 & 4\\
LRFR \cite{LRFR} &53.4 & 73.0 &80.6& 0.852     & 0.750     & 0.683 & 4\\ 
QDFL \cite{QDFL} & 56.5 & 79.8 &86.7& 0.906     & 0.825     & 0.759  & 20 \\
DAC \cite{DAC} & 58.3 & 78.6 & 84.4 & 0.899     & 0.808     & 0.733 & 8\\
MCCG \cite{shen2023mccg} & 56.7 & \underline{80.5} & \textbf{88.6 } & 0.901     & \textbf{0.841}     & \textbf{0.779}   & 6\\
Sample4Geo \cite{s4g} & \underline{58.6}& 79.9 & 86.4  & \underline{0.907}     & 0.825     & 0.760 & 13\\
CAMP \cite{wu2024camp} & \textbf{62.4} & \textbf{82.7} & \underline{88.3} & \textbf{0.920}     & \underline{0.834}     & \underline{0.766} & 6\\

\hline\hline
\end{tabular}
}

\label{tab:retrieval_result}
\end{table}

\begin{table*}[!ht]
\centering
\caption{\textbf{Localization Results of Different Image Retrieval Approaches.}} 
\scalebox{0.78}{
\begin{tabular}{l |c c c |c c c |c c c |c c c }
\hline\hline
\textbf{Method} & \multicolumn{3}{c|}{\textbf{Top1 + SP\_LG}} & \multicolumn{3}{c|}{\textbf{Top1 + RoMa}} & \multicolumn{3}{c|}{\textbf{Top5 Re-rank + SP\_LG}} &  \multicolumn{3}{c}{\textbf{Top5 Re-rank + RoMa}} \\
\cmidrule(lr){2-4} \cmidrule(lr){5-7} \cmidrule(lr){8-10} \cmidrule(lr){11-13} 
& \textbf{A@5m} & \textbf{A@10m} & \textbf{A@20m} & \textbf{A@5m} & \textbf{A@10m} & \textbf{A@20m} & \textbf{A@5m} & \textbf{A@10m} & \textbf{A@20m} & \textbf{A@5m} & \textbf{A@10m }& \textbf{A@20m} \\
\hline\hline

NCC \cite{NCC} & 7.5 & 9.8 & 11.4 & 10.8 & 13.3 & 15.2 & 16.5 & 22.1 & 26.2 & 22.4 & 28.3 & 31.4 \\
MI \cite{MI} & 13.6 & 18.6 & 21.4 & 18.3 & 21.7 & 25.4 & 28.9 & 38.8 & 45.0 & 36.4 & 44.6 & 49.7 \\
\addlinespace[0.3em]
\hdashline
\addlinespace[0.3em]
NetVLAD \cite{netvlad} & 41.7 & 55.3 & 62.8 & 49.3 & 60.0 & 65.1 & 55.0 & 72.1 & 80.7 & 64.3 & 76.6 & 82.9 \\
LPN \cite{LPN} & 43.6 & 57.3 & 66.5 & 54.1 & 64.2 & 69.5 & 57.1 & 75.0 & 84.6 & 66.8 & 80.5 & 86.4 \\
RK-Net \cite{RK-Net} & 49.9 & 66.6 & 76.2 & 62.2 & 73.9 & 79.4 & 60.0 & 77.5 & 87.2 & 68.6 & 82.6 & 89.1 \\
FSRA \cite{FSRA} & 51.6 & 69.3 & 79.4 & 64.9 & 76.7 & 82.2 & 60.7 & 80.5 & 90.4 & 71.7 & 84.9 & 91.7 \\
DenseUAV \cite{DenseUAV} & 49.7 & 66.5 & 76.4 & 61.7 & 72.9 & 79.8 & 59.0 & 78.2 & 88.1 & 70.3 & 84.2 & 90.6 \\

MEAN \cite{MEAN} & 48.8 & 65.4 & 72.9 & 59.8 & 70.6 & 76.8&  59.0 &  79.3 &  88.1 &  70.2 &  84.1 &  90.6 \\
LRFR \cite{LRFR} & 51.6  & 68.8  & 77.2 & 63.1 & 73.4 & 79.9 & 60.0 & 79.5 & 88.6 & 71.4 & 84.8 & 91.5\\ 

QDFL \cite{QDFL} & 55.1 & 74.4 & 82.7 & 67.5 & 79.4 & 85.8 & \underline{62.4} & 81.5 &  91.1 & 73.0 & 86.2 &  93.0 \\
DAC \cite{DAC} & 53.4 & 71.3 & 81.7 & 66.9 & 78.8 & 85.1 & 62.2 & 81.8 & 91.6 & 72.7 & 86.9 & 93.0 \\
MCCG \cite{shen2023mccg} & \underline{54.9} & \underline{74.2} &\underline{ 83.6} & \underline{68.2} & \textbf{81.4} & \underline{87.4} & \textbf{63.1} & 82.1 & 91.9 & 72.7 & 86.2 & 92.8 \\
Sample4Geo \cite{s4g} & 54.1 & 72.8 & 83.2 & 66.9 & 78.8 & 85.7 & \underline{62.4} & \textbf{82.4} & \underline{91.9} & \underline{73.2} & \underline{87.5} & \underline{93.7} \\
CAMP \cite{wu2024camp} & \textbf{55.8} & \textbf{75.0} & \textbf{85.1} & \textbf{70.1} & \underline{81.3} & \textbf{87.6} & \underline{62.4} & \underline{82.2}& \textbf{92.4} & \textbf{74.6} & \textbf{87.6} & \textbf{94.2} \\

\hline\hline
\end{tabular}
}
\label{tab:retrieval_AVL}
\end{table*}

This section aims to identify the best retrieval approaches.
As shown in \cref{tab:retrieval_result}, the handcrafted template matching approaches, NCC and MI, perform worst since they are sensitive to viewpoint difference. 

Among deep learning methods, models like CAMP, Sample4Geo, DAC, and  MCCG which use ConvNeXt \cite{liu2022convnet} as the backbone achieved better performance: MCCG optimizes network structure for better feature representation, while Sample4Geo, DAC and CAMP focus on the improvements of training and sample strategies, which reveals that training and sampling strategies are more effective for improving the accuracy of current image retrieval models.

Although DenseUAV proposed a new baseline for AVL, it underperformed in multi-view conditions, possibly because the model was mainly designed for nadir-view images. Overall, drone-to-ortho reference map retrieval
under low-altitude, multi-view conditions is still challenging.

To analyze the impact of retrieval accuracy on the subsequent localization accuracy, we also combined retrieval approaches with different matching approaches (SP+LG and RoMa) and different localization strategies (Top1 Matching and Top5 Re-rank based on the number of inliers). 

From \cref{tab:retrieval_AVL}, we can observe that CAMP achieves the highest localization accuracy. This is consistent with the trend observed in \cref{tab:retrieval_result}.

\subsection{Image Retrieval Metric}
\label{sec:metric}
\begin{figure*}[!h]
    \centering
    \includegraphics[width=1\textwidth]{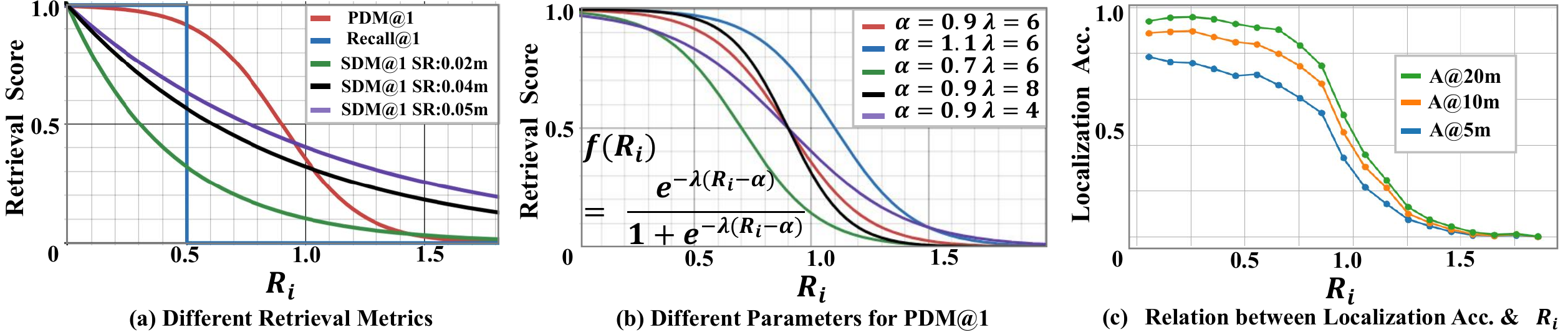}
    \caption{\textbf{Illustration of PDM@K.} (a) Different retrieval metric comparison. For clarity in the same figure, we have converted the spatial distance $d_i$ of SDM@1 to $R_i$ and the threshold for Recall@1 is set to 0.5. (b) Different parameter combinations for PDM@1. (c) Relation between localization accuracy and $R_i$. This curve is based on the actual AVL experiment results.}
    \label{fig:PDE_expalin}
\end{figure*}

\begin{figure}[!h]
    \centering
    \includegraphics[width=0.47\textwidth]{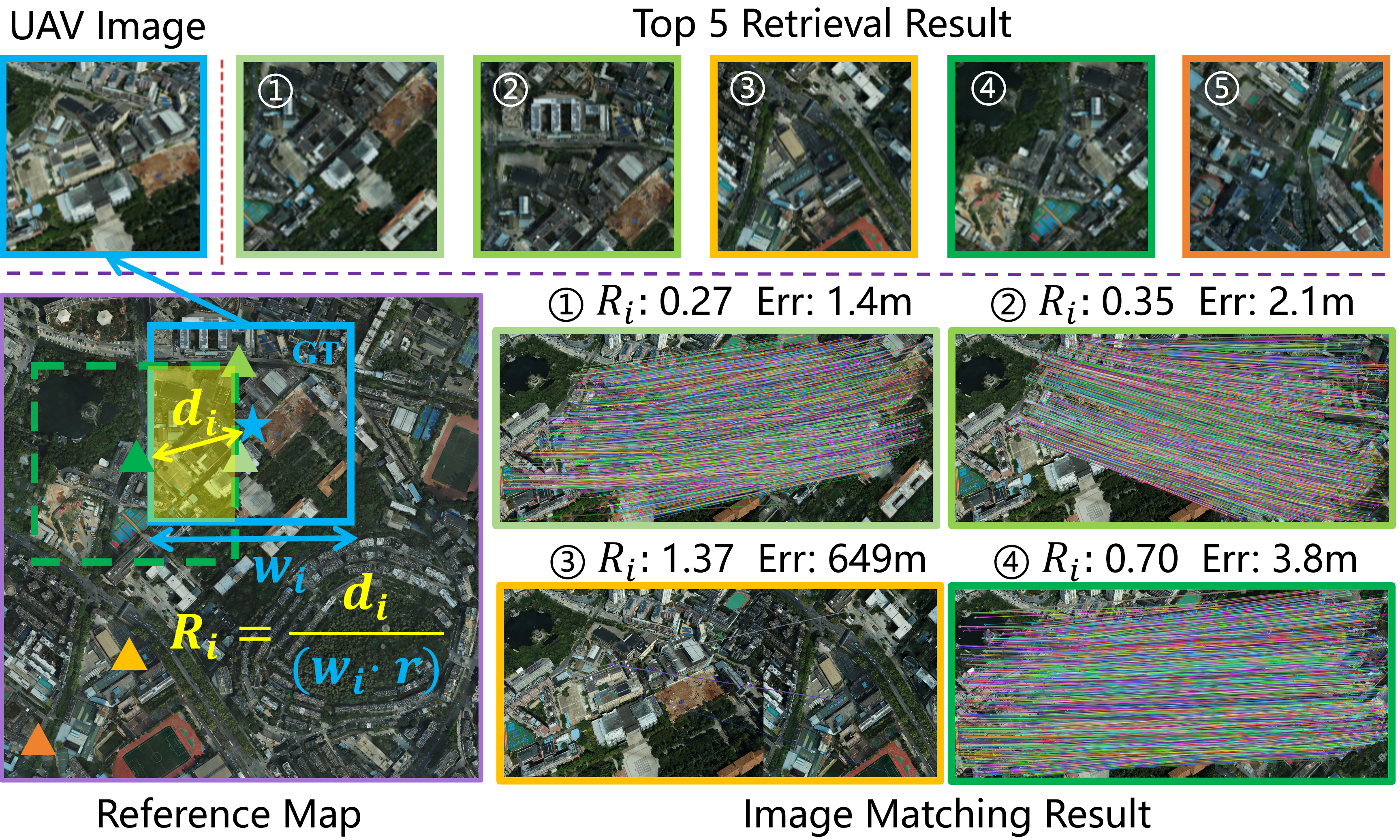}
    \caption{\textbf{Visualization of the Relation between $R_i$ in PDM@K and Subsequent Localization Accuracy. }}
    \label{fig:PDE_visual}
\end{figure}

In UAV AVL tasks, reference maps are geographically continuous, yet Recall@K solely focuses on the result with minimal localization error, disregarding the distribution of other images within Top K. As shown in \cref{fig:PDE_expalin}a, Recall@1 assigns a binary score (1 or 0) to retrieval results, which fails to account for the spatial distribution of retrievals. For instance, in \cref{fig:PDE_visual}, the 4th ranked image, despite its distance from the ground truth, effectively matches the UAV image with a localization error of 3.8m. Therefore, it is unreasonable for Recall@K to assign this result a score of 0.


The Spatial Distance Metric \cite{DenseUAV} (SDM@K, see \cref{equ:SDK}) partially addresses this issue by considering the spatial distribution of retrieval results. However, SDM@K relies on the spatial distance 
 $d_i$ between the ground truth and the closest retrievals, making it sensitive to variations in the spatial resolution of reference maps.  This leads to inconsistent score curves across different maps (see \cref{fig:PDE_expalin}a), which is unfair for evaluation.
\begin{equation}
\resizebox{.8\hsize}{!}{$ SDM@K = \sum\limits_{i = 1}^K {\frac{{(K - i + 1)}}{{{e^{s \cdot {d_i}}}}}} /\sum\limits_{i = 1}^K {(K - i + 1)} .$}
    \label{equ:SDK}
\end{equation}

In contrast, PDM@K normalizes both the image size and spatial resolution with $R_i=d_i/(w_i\cdot r)$, ensuring a fair evaluation across diverse regions' reference maps. 

Moreover, PDM@K effectively bridges image retrieval accuracy with final localization accuracy. As illustrated in \cref{fig:PDE_visual}, $R_i$ is correlated with the distance between the retrieved position and ground truth, reflecting the overlap rate of image pairs for subsequent image matching-based localization. Higher $R_i$ indicates lower overlap, increasing image matching difficulty and reducing localization accuracy. For example, in \cref{fig:PDE_visual}, the first retrieval with $R_i = 0.27$ achieves an error of 1.4m, while the second retrieval with $R_i = 0.35$ has an error of 2.1m. In contrast, the third retrieval with $R_i = 1.37$ leads to localization failure, with an error of 649m due to insufficient overlap. This trend is further supported by the experimental localization accuracy distribution presented in \cref{fig:PDE_expalin}c.

For the retrieval score function $f(R_i)$  (see \cref{fig:PDE_expalin}b), $\alpha$ determines the $R_i$ threshold at which the score drops, while $\lambda$ controls the sharpness of the decay. When $R_i$ exceeds the normalized diagonal length of the image, $l$ (e.g., $l=1.67$ when the aspect ratio of drone image is 4:3), there is no overlap between the images, resulting in a score of 0. Therefore, $\alpha$ should be set slightly above $l/2$, and $\lambda$ can be adjusted based on the impact of different overlap rates on matching accuracy. On the AnyVisLoc dataset, $\lambda$ is recommended to be in the range of 4 to 8.

\subsection{Image Matching Approach}
\label{sec:matching}
To evaluate various image matching approaches, we conducted tests using CAMP and 19 matching methods.  As shown in \cref{tab:matching_result}, learning-based models generally outperform traditional handcrafted methods like SIFT and ORB. While ORB is fast and widely used in SLAM \cite{ORB-SLAM3}, it struggles with viewpoint changes. Additionally, RoMa performs best among dense matching methods, whereas SP+LG$_{GIM}$+k2s leads in sparse matching. Overall, dense methods significantly surpass sparse methods in localization accuracy, but at the cost of lower computational efficiency (e.g., RoMa takes over six times longer than SP+LG$_{GIM}$+k2s). Although DeDoDe attempts to address this issue, it has not achieved good trade-off in our test. Consequently, sparse matching approaches remain more advantageous for real-time localization tasks.

\begin{table*}[h!]
\centering
\caption{Localization Results of Different Image Matching Approaches. $\dag$ denotes the approaches that run on CPU.}
\scalebox{0.75}{
\begin{tabular}{ l |c c c c c|c c c c c|c }
\hline\hline
\textbf{Image Matching Method} & \multicolumn{5}{c|}{\textbf{Top1 + CAMP}} & \multicolumn{5}{c|}{\textbf{Top5 Re-rank + CAMP}} & \textbf{Run Time}\\
\cmidrule(lr){2-6} \cmidrule(lr){7-11}   & \textbf{A@5m} & \textbf{A@7m} & \textbf{A@10m} & \textbf{A@15m} & \textbf{A@20m}   & \textbf{A@5m} & \textbf{A@7m} & \textbf{A@10m} & \textbf{A@15m} & \textbf{A@20m}   & \textbf{ms/frame}\\ \hline\hline
SIFT \cite{SIFT}  & 43.9 & 50.1 & 55.7 & 59.9 & 62.5   & 52.5 & 59.4 & 64.5 & 68.4 & 70.8    & 316\dag\\
ORB \cite{ORB-SLAM3}   & 3.9 & 5.8 & 8.0 & 10.9 & 13.6    & 5.4 & 7.6 & 10.3 & 13.8 & 17.2    & 44\dag\\
\addlinespace[0.3em] 
 \hdashline
 \addlinespace[0.3em]
 D2Net \cite{D2-Net}  & 27.3 & 38.1 & 51.6 & 61.9 & 68.6   & 33.0& 44.3 & 59.8 & 72.2 & 78.5   & 4083\\
ALIKE \cite{alike}   & 27.7 & 32.0 & 34.8 & 38.2 & 39.6   & 31.9 & 37.3 & 40.8 & 44.4 & 45.9   & 268\\
Xfeat* \cite{potje2024xfeat}  & 35.2 & 44.5 & 54.4 & 62.1 & 66.8    & 42.8 & 53.4 & 63.5 & 70.7 & 75.4    & 69\\
LiftFeat \cite{LiftFeat} & 30.3 & 38.9 & 47.3 & 55.2 & 59.1 & 39.3 & 50.7 & 60.6 & 69.4 & 73.5 & 40\\
SP \cite{detone18superpoint}+SG \cite{superglue}  & 52.1 & 63.3 & 71.7 & 77.8 & 80.9  & 59.1 & 71.3 & 79.6 & 85.8 & 89.6   &  92 \\
SP \cite{detone18superpoint}+SG \cite{superglue}+Omni \cite{omniglue}  & 45.3 & 55.0 & 64.2 & 72.3 & 76.1  & 52.2 & 63.6 & 72.9 & 81.3 & 84.8    & 3116\dag \\
SP \cite{detone18superpoint}+LG \cite{lightglue} & 46.6 & 55.8 & 63.5 & 68.7 & 71.7 & 55.7 & 66.5 & 76.2 & 81.9 & 85.4 & 76\\
SP \cite{detone18superpoint}+LG$_{MINIMA}$ \cite{lightglue, ren2025minima} & 46.0 & 55.6 & 63.1 & 68.4 & 72.4 & 55.2 & 65.8 & 75.0 & 81.7 & 85.8 & 74\\
SP \cite{detone18superpoint}+LG$_{GIM}$ \cite{lightglue, xuelun2024gim} & 55.8 & 66.7 & 75.0 & 81.6 & 85.1   & 62.4 & 73.5 & 82.2 & 88.9 & \underline{92.4}   &  75 \\
SP \cite{detone18superpoint}+LG$_{GIM}$ \cite{lightglue, xuelun2024gim}+k2s \cite{kim2024keypt2subpx}  & 57.0 & 67.6 & 75.4 & 81.7 & 84.8  & 62.9 & 74.8 & 83.2 & 89.2 & 91.8   & 105\\
\addlinespace[0.3em] 
 \hdashline
 \addlinespace[0.3em] 
XoFTR$_{MINIMA}$ \cite{tuzcuouglu2024xoftr, ren2025minima} & 48.3 & 54.7 & 60.2 & 63.4 & 65.8 & 59.4 & 66.8 & 72.9 & 77.4 & 80.0 & 68\\ 
LoFTR$_{GIM}$ \cite{sun2021loftr, xuelun2024gim} & 59.5 & 67.4 & 74.1 & 79.1 & 81.6   & 65.9 & 74.9 & 82.5 & 87.2 & 89.1 & 165\\
DeDoDe \cite{dedode}  & 51.5 & 61.8 & 69.1 & 74.8 & 77.2   & 61.3 & 71.6 & 79.0 & 84.4 & 86.9   & 291\\

DKM$_{GIM}$ \cite{DKM, xuelun2024gim}  & \underline{65.6} & \underline{73.8} & \underline{80.2} & \underline{84.0} & \underline{86.4}   & 68.5 & \underline{78.1} & \underline{85.0} & \underline{90.3} & 92.1  & 4915\\
RoMa$_{GIM}$ \cite{roma, xuelun2024gim} & 60.3 & 66.5 & 71.4 & 74.7 & 76.2 & 68.2 & 74.6 & 81.7 & 86.7 & 88.6 & 655\\
RoMa$_{MINIMA}$ \cite{roma, ren2025minima} & 60.0 & 65.2 & 69.8 & 74.0 & 76.0 & \underline{69.8} & 76.8 & 82.5 & 86.4 & 88.5 & 653\\
RoMa \cite{roma}   & \textbf{70.1} & \textbf{76.1} & \textbf{81.3} & \textbf{85.7} & \textbf{87.6}  & \textbf{73.9} & \textbf{80.8} & \textbf{87.2} & \textbf{91.5} &\textbf{94.0}  & 659\\

\hline\hline
\end{tabular}
}

\label{tab:matching_result}
\end{table*}


\subsection{Localization Strategy}
\label{sec:strategy}
\begin{table}[h!]
\centering
\caption{\textbf{Performance Metrics of Localization Strategies}}
\scalebox{0.66}{
\begin{tabular}{l|c c c| c c}
\hline\hline
\textbf{Strategy } & \textbf{A@5m} & \textbf{A@10m} & \textbf{A@20m} & \textbf{s/frame} &  \textbf{storage}\\
\hline\hline
Matching-wo-Retrieval \cite{yuan2020automated} & 34.3 & 45.7 & 54.3  & 1.4 & Large\\
Top1 Matching \cite{ye2024coarse} & 55.8 & 74.3 & 84.0  & \textbf{0.3} & medium\\
Top5 Re-rank (N=5) \cite{chen2021real} & \underline{62.2} & \underline{82.4 }& \underline{91.5}  & \underline{0.8} & medium\\
Most Inliers \cite{gurgu2022vision} & \textbf{64.0} & \textbf{83.2} & \textbf{92.6}  & 10.2 & small\\
\hline\hline
\end{tabular}
}
\label{tab:strategys}
\end{table}
\begin{table}[h!]
	\centering
 	\caption{\textbf{Performance Metrics for Different Reference Maps}}
 \scalebox{0.62}{
	\begin{tabular}{c|cc|cc|ccc}
 \hline\hline
		\textbf{Map}       & \multicolumn{2}{c|}{\textbf{Spatial Resolution (Avg.)}}    & \multicolumn{2}{c|}{\textbf{Retrieval Acc.}}     & \multicolumn{3}{c}{\textbf{Localization Acc.}} \\
  \cmidrule(lr){2-4} \cmidrule(lr){4-5}     \cmidrule(lr){6-8}    
	& \textbf{2D-Reference}   & \textbf{DSM}    & \textbf{R@1 }  & \textbf{P@1}   & \textbf{A@5m } & \textbf{A@10m} & \textbf{A@20m }    \\ \hline\hline
 Aerial   & 0.070m   & 0.947m &  \textbf{61.6} & \textbf{0.922} &  \textbf{74.1}       & \textbf{87.7 }      & \textbf{94.2}        \\
	Satellite & 0.197m   & 30m   &  42.8  & 0.814  & 18.5   & 38.7  & 58.5    \\
    
		  \hline\hline    
	\end{tabular}
 }
	\label{tab:reference}
\end{table}

This section compares four distinct localization strategies, the performance metrics are shown in \cref{tab:strategys}. The retrieval method employed was CAMP, and the matching method was SP+LG$_{GIM}$+k2s.

The \textbf{Direct Matching without Retrieval} strategy performed the worst. In UAV AVL tasks, the area of the reference map (e.g., $9km^2$) is often much larger than the area of the UAV image (e.g., $0.04km^2$).  This discrepancy leads to redundant areas in the reference map, which greatly expands the search space of the matching algorithm and result in poor localization robustness.

The \textbf{Top1 Matching} strategy uses image retrieval to narrow the search space for subsequent image matching. However, due to the insufficient accuracy of existing image retrieval networks, when the Top1 image is not the ground truth, the subsequent matching algorithm will fail.

The \textbf{Top N Re-rank} strategy re-ranks the Top N retrievals based on the number of inliers. The basic assumption is that images with more inliers are often closer to the ground truth and are more suitable for PnP problem solving. Compared to the Top1 strategy, this strategy significantly enhances accuracy with an acceptable increase in time.

The \textbf{Most Inliers} strategy directly matches the UAV image with all gallery images and also uses the number of inliers to sort retrievals. Although it has slightly higher localization accuracy than the Top N Re-Rank strategy, feature matching for each gallery image is very time-consuming, making it unable to meet real-time requirements.

Overall, \textbf{Top N Re-rank} strategy has a comprehensive advantage in localization accuracy, computation time, and memory usage. 

\subsection{The Best Combined Method: Baseline}
\label{sec:baseline}
Based on the above experiments, the best combined method which consists of \textbf{CAMP+RoMa+Top N Rerank} was chosen as the baseline. This baseline was used to test all data in the AnyVisLoc dataset, with results shown in \cref{tab:reference}. Although the baseline showed good accuracy at 94.2\% within 20 meters (on the aerial map), it only achieved \textbf{74.1\%} accuracy within 5 meters, which may not meet the precise AVL requirements under low-altitude, multi-view conditions.

\section{Factors Affecting the Baseline Accuracy }
This section discussed the main factors influencing the localization performance of current baseline. The full dataset is used to conduct the experiment.

\begin{figure}[!h]
    \centering
    \includegraphics[width=0.48\textwidth]{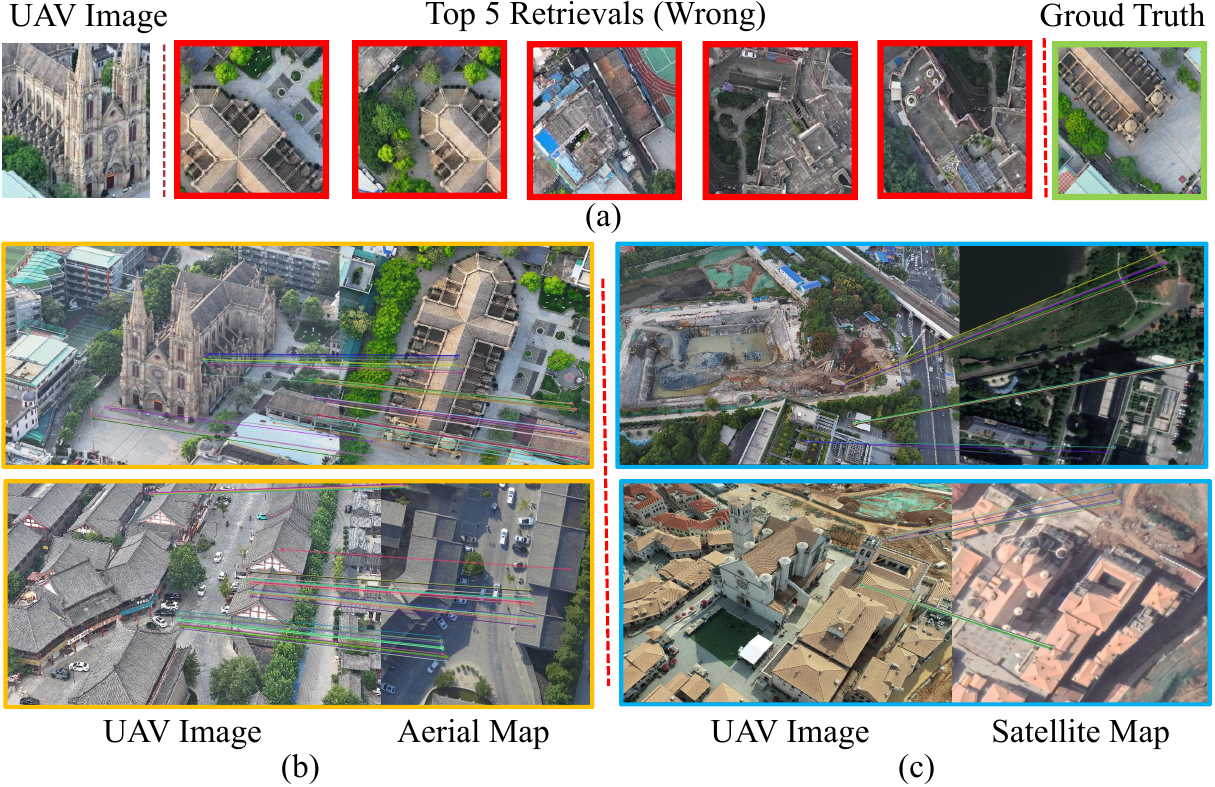}
    \caption{\textbf{Challenging Cases for UAV AVL.} The low-altitude multi-view conditions present greater challenge for image retrieval (a) and image matching (b). Temporal and modality differences in satellite maps makes localization more difficult (c).}
    \label{fig:failed_case}
\end{figure}

\subsection{Reference Map}
Different reference maps exhibit different characteristics and localization accuracy (see  \cref{tab:reference}). \textbf{Satellite maps} are easier to obtain but significantly less accurate for localization. This is mainly caused by two factors: Firstly, satellite maps have lower spatial resolution, especially with DSM data at 30m. Under low-altitude multi-view conditions, substantial elevation differences in UAV images (e.g., rooftops vs. flat ground) require high-resolution DSM data (e.g. $<1m$) to ensure accurate PnP problem solving. Secondly, the significant temporal and modality differences between UAV images and satellite maps pose greater challenges for image retrieval and matching (see \cref{fig:failed_case}c). 

\textbf{Aerial maps} provide superior localization accuracy, however, pre-aerial photography and precise 3D modeling of the flight area are required, making them less suitable for  time-sensitive missions (e.g., emergency rescue) or long-distance flight tasks. Consequently, the choice of reference map type should be tailored to the specific requirements of the mission at hand.

To better analyze the impact of other factors, the following experiments are conducted based on the aerial map. 

\subsection{Multi-view Observation}
\begin{figure}[!h]
    \centering
    \includegraphics[width=0.3\textwidth]{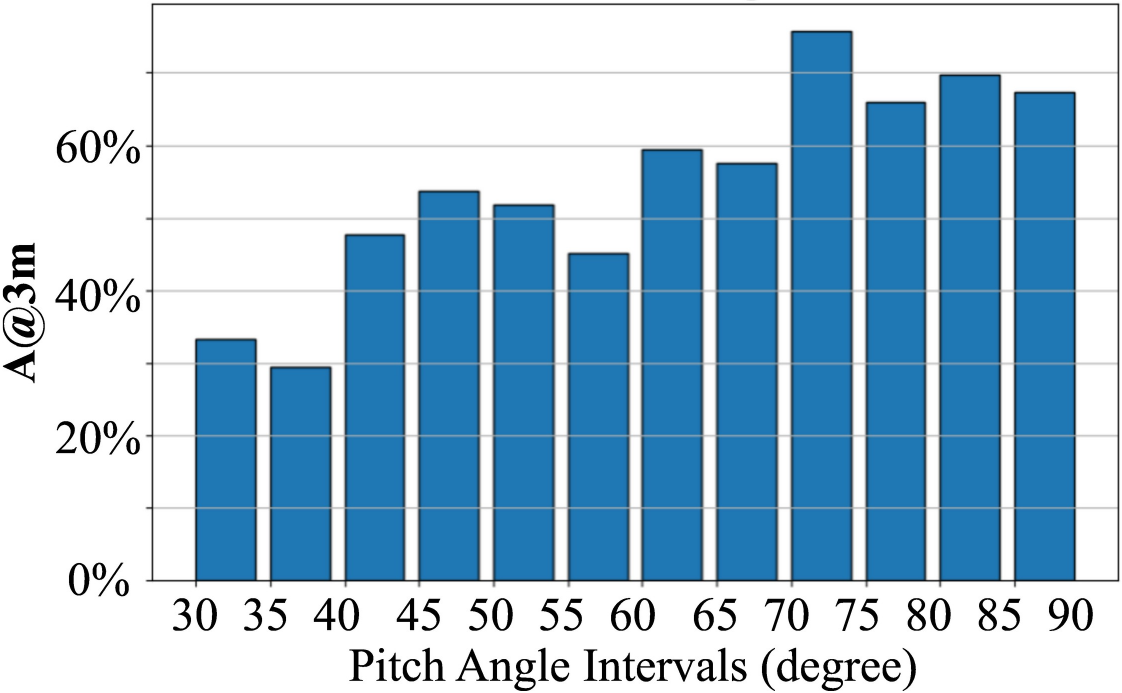}
    \caption{\textbf{Impact of Pitch Angle on Localization Accuracy.} A smaller pitch angle tends to reduce localization accuracy.}
    \label{fig:pitch_curve}
\end{figure}
This section analyzes the localization results  across different pitch angles. As shown in \cref{fig:pitch_curve},  a smaller pitch angle tends to reduce localization accuracy, indicating that oblique-view images present greater challenges for accurate localization compared to nadir-view images. In low-altitude flights, oblique-view images capture substantial side-view information from 3D objects. These side-view details reduce
 both global and local similarity between UAV images and the ortho-reference map. Consequently, the accuracy of image retrieval and matching algorithms is diminished (see \cref{fig:failed_case}), making precise localization more difficult.

\subsection{Noise in Prior Information}
\label{sec:prior_inf}
\begin{table}
\centering
\caption{\textbf{Impact of Different Prior Information Noise on the Localization Accuracy. }Std: standard deviation.}
\label{tab:auc_values}
\begin{subtable}[h]{0.44\linewidth}
\scalebox{0.52}{
\begin{tabular}{c|ccc}
\hline\hline
\textbf{Yaw Std} & \textbf{A@5m} & \textbf{A@10m} & \textbf{A@20m} \\
\hline\hline
0° & 74.6 & 87.6 & 94.2 \\
5° & 74.3 ($\downarrow 0.3$) & 86.3 & 93.2 \\
10° & 72.7 ($\downarrow 1.9$) & 86.4 & 92.8 \\
20° & 72.4 ($\downarrow 2.2$) & 86.4 & 93.6 \\
30° & 70.5 ($\downarrow \textbf{4.1}$) & 83.5 & 90.2 \\
50° & 60.9 ($\downarrow 13.7$) & 72.8 & 79.5 \\
60° & 48.9 ($\downarrow \textbf{25.7}$) & 63.0 & 70.0 \\
\hline\hline
\end{tabular}}
\caption{Yaw Noise}
\label{tab:auc_values.b}
\end{subtable}
\hfill
\begin{subtable}[h]{0.44\linewidth}
\scalebox{0.52}{
\begin{tabular}{c|ccc}
\hline\hline
\textbf{Pitch Std} & \textbf{A@5m} & \textbf{A@10m} & \textbf{A@20m} \\
\hline\hline
0° & 74.6 & 87.6 & 94.2 \\
3° & 73.9 ($\downarrow 0.7$) & 87.2 & 94.2 \\
5° & 73.9 ($\downarrow \textbf{0.7}$) & 87.1 & 93.2 \\
7° & 73.3 ($\downarrow 1.3$) & 87.3 & 93.4 \\
10° & 72.1 ($\downarrow 2.5$) & 86.5 & 92.2 \\
20° & 71.6 ($\downarrow 3.0$) & 86.2 & 92.9 \\
30° & 69.7 ($\downarrow \textbf{4.9}$) & 83.7 & 90.4 \\
\hline\hline
\end{tabular}}
\caption{Pitch Noise}
\label{tab:auc_values.a}
\end{subtable}
\label{tab:prior_noise}
\end{table}

Current UAV platforms generally carry sensors that can provide information about pitch/yaw angles and altitude (e.g., gyroscopes and altimeters). In this paper, this information is used to align the UAV image and the reference map to similar rotation and scale, reducing the search space for retrieval and matching.  However, in the real scenarios, this prior information often contains noise. To analyze this issue, we added different levels of Gaussian noise (mean of 0) to the prior information. The results are shown in \cref{tab:prior_noise}.

\textbf{Yaw noise} affects the rotation estimation. When $std > 10^\circ$, the accuracy of existing image retrieval and matching methods is significantly affected. Notably, the larger the std, the more severely the localization accuracy declines. When the std increases to 30° and 60°, A@5m decreases by 4.1\% and 25.7\%, respectively. 

\textbf{Pitch noise} affects the scale estimation in \cref{equ:scale_estimate}. Low pitch noise ($std < 5^\circ$) has minimal impact on localization accuracy. However, as pitch noise increases ($std > 7^\circ$), the significant scale difference complicates image retrieval and matching, leading to a noticeable drop in accuracy, with A@5m decreasing by 4.9\% at $std = 30^\circ$. 
\begin{equation}
  r = \frac{{altitude}}{{\sin (pitch)}} \cdot \tan (\frac{{FOV}}{2}) \cdot \frac{2}{{\sqrt {({w^2} + {h^2})} }.}  
  \label{equ:scale_estimate}
\end{equation}

\textbf{Altitude noise} also affects scale estimation like pitch, but its impact is small, the details are put in the appendix.

\section{Conclusion}
This paper benchmarks UAV AVL under low-altitude multi-view conditions. The AnyVisLoc dataset is constructed to facilitate comprehensive evaluation. A unified framework is introduced to integrate image-level retrieval, pixel-level matching methods, and various localization strategies. Based on this dataset and framework, state-of-the-art AVL approaches are exhaustively evaluated to select the best combined method as the baseline. Additionally, key factors affecting the baseline performance are thoroughly discussed. Overall, the baseline method achieved a 74.1\% accuracy within 5 meters, indicating substantial room for improvement and highlighting the potential of our benchmark to drive future research in this field.


{
    \small
    \bibliographystyle{ieeenat_fullname}
    \bibliography{main}

@String(CVPR= {IEEE Conf. Comput. Vis. Pattern Recog.})

@String(ICCV= {Int. Conf. Comput. Vis.})

@String(ECCV= {Eur. Conf. Comput. Vis.})

@String(TIP  = {IEEE Trans. Image Process.})

@String(ICLR = {Int. Conf. Learn. Represent.})

@String(AAAI = {AAAI})

@String(CVPRW= {IEEE Conf. Comput. Vis. Pattern Recog. Worksh.})

@String(CVPR  = {CVPR})

@String(ICCV  = {ICCV})

@String(ECCV  = {ECCV})

@String(TIP   = {IEEE TIP})

@String(TCSVT = {IEEE TCSVT})

@String(ICLR  = {ICLR})

@String(CVPRW= {CVPRW})

@inproceedings{GPS_jamming,
author = {Xue, Nian and Niu, Liang and Hong, Xianbin and Li, Zhen and Hoffaeller, Larissa and P\"{o}pper, Christina},
title = {DeepSIM: GPS Spoofing Detection on UAVs using Satellite Imagery Matching},
year = {2020},
publisher = {Association for Computing Machinery},
address = {New York, NY, USA},
url = {https://doi.org/10.1145/3427228.3427254},
doi = {10.1145/3427228.3427254},
booktitle = {Proceedings of the 36th Annual Computer Security Applications Conference},
pages = {304–319},
numpages = {16},
location = {Austin, USA},
series = {ACSAC '20}
}

@inproceedings{
xuelun2024gim,
title={GIM: Learning Generalizable Image Matcher From Internet Videos},
author={Xuelun Shen and Zhipeng Cai and Wei Yin and Matthias Müller and Zijun Li and Kaixuan Wang and Xiaozhi Chen and Cheng Wang},
booktitle={ICLR},
year={2024}
}

@InProceedings{kim2024keypt2subpx,
    author = {Shinjeong Kim and Marc Pollefeys and Daniel Barath},
    title = {Learning to Make Keypoints Sub-Pixel Accurate},
    booktitle = {ECCV},
    year = {2024}
}

@inproceedings{detone18superpoint,
  author    = {Daniel DeTone and
               Tomasz Malisiewicz and
               Andrew Rabinovich},
  title     = {SuperPoint: Self-Supervised Interest Point Detection and Description},
  booktitle = {CVPRw},
  year      = {2018},
}

@inproceedings{lightglue,
  author    = {Philipp Lindenberger and
               Paul-Edouard Sarlin and
               Marc Pollefeys},
  title     = {{LightGlue: Local Feature Matching at Light Speed}},
  booktitle = {ICCV},
  year      = {2023}
}

@ARTICLE{SIFT,
  author={D. G. Lowe},
  journal={Int. J. Comput. Vis.}, 
  title={Distinctive image features from scale-invariant keypoints}, 
  year={2004},
  volume={60},
  number = {2},
  pages={91–110},
}

@inproceedings{ren2025minima,
  title={{MINIMA}: Modality invariant image matching},
  author={Ren, Jiangwei and Jiang, Xingyu and Li, Zizhuo and Liang, Dingkang and Zhou, Xin and Bai, Xiang},
  booktitle={Proceedings of the IEEE/CVF Conference on Computer Vision and Pattern Recognition},
  pages={23059--23068},
  year={2025},
  publisher ={Piscataway: IEEE}
}

@article{QDFL,
  title={Query-Driven Feature Learning for Cross-View Geo-Localization},
  author={Hu, Shuyu and Shi, Zelin and Jin, Tong and Liu, Yunpeng},
  journal={IEEE Transactions on Geoscience and Remote Sensing},
  year={2025},
  publisher={IEEE}
}

@article{MEAN,
  title={Multi-level embedding and alignment network with consistency and invariance learning for cross-view geo-localization},
  author={Chen, Zhongwei and Yang, Zhao-Xu and Rong, Hai-Jun},
  journal={IEEE Transactions on Geoscience and Remote Sensing},
  year={2025},
  publisher={IEEE}
}

@article{LRFR,
  title={Learning Robust Feature Representation for Cross-View Image Geo-localization},
  author={Gan, Wenjian and Zhou, Yang and Hu, Xiaofei and Zhao, Luying and Huang, Gaoshuang and Hou, Mingbo},
  journal={IEEE Geoscience and Remote Sensing Letters},
  year={2025},
  publisher={IEEE}
}

@article{nikolakopoulos2020accuracy,
  title={Accuracy assessment of ALOS AW3D30 DSM and comparison to ALOS PRISM DSM created with classical photogrammetric techniques},
  author={Nikolakopoulos, Konstantinos G},
  journal={European Journal of Remote Sensing},
  volume={53},
  number={sup2},
  pages={39--52},
  year={2020},
  publisher={Taylor \& Francis}
}

@article{li2022global,
  title={Global DEMs vary from one to another: an evaluation of newly released Copernicus, NASA and AW3D30 DEM on selected terrains of China using ICESat-2 altimetry data},
  author={Li, Hui and Zhao, Jiayang and Yan, Bingqi and Yue, Linwei and Wang, Lunche},
  journal={International Journal of Digital Earth},
  volume={15},
  number={1},
  pages={1149--1168},
  year={2022},
  publisher={Taylor \& Francis}
}

@inproceedings{tuzcuouglu2024xoftr,
  title={XoFTR: Cross-modal Feature Matching Transformer},
  author={Tuzcuo{\u{g}}lu, {\"O}nder and K{\"o}ksal, Aybora and Sofu, Bu{\u{g}}ra and Kalkan, Sinan and Alatan, A Aydin},
  booktitle={Proceedings of the IEEE/CVF Conference on Computer Vision and Pattern Recognition},
  pages={4275--4286},
  year={2024}
}

@INPROCEEDINGS{LiftFeat,
  author={Liu, Yepeng and Lai, Wenpeng and Zhao, Zhou and Xiong, Yuxuan and Zhu, Jinchi and Cheng, Jun and Xu, Yongchao},
  booktitle={2025 IEEE International Conference on Robotics and Automation (ICRA)}, 
  title={LiftFeat: 3D Geometry-Aware Local Feature Matching}, 
  year={2025},
  volume={},
  number={},
  pages={11714--11720},
  doi={10.1109/ICRA55743.2025.11127853}
}

@article{zhou2025cdm,
  title={CDM-Net: A Framework for Cross-View Geo-Localization with Multimodal Data},
  author={Zhou, Xin and Yang, Xuerong and Zhang, Yanchun},
  journal={IEEE Transactions on Geoscience and Remote Sensing},
  year={2025},
  publisher={IEEE}
}

@inproceedings{superglue,
  author    = {Paul-Edouard Sarlin and
               Daniel DeTone and
               Tomasz Malisiewicz and
               Andrew Rabinovich},
  title     = {{SuperGlue}: Learning Feature Matching with Graph Neural Networks},
  booktitle = {CVPR},
  year      = {2020},
  url       = {https://arxiv.org/abs/1911.11763}
}

@inproceedings{NCC,
  title={Absolute localization using image alignment and particle filtering},
  author={Van Dalen, Gerald J and Magree, Daniel P and Johnson, Eric N},
  booktitle={AIAA Guidance, Navigation, and Control Conference},
  pages={0647},
  year={2016}
}

@inproceedings{netvlad,
  title={NetVLAD: CNN architecture for weakly supervised place recognition},
  author={Arandjelovic, Relja and Gronat, Petr and Torii, Akihiko and Pajdla, Tomas and Sivic, Josef},
  booktitle={CVPR},
  pages={5297--5307},
  year={2016}
}

@inproceedings{wang2024efficient,
  title={Efficient LoFTR: Semi-dense local feature matching with sparse-like speed},
  author={Wang, Yifan and He, Xingyi and Peng, Sida and Tan, Dongli and Zhou, Xiaowei},
  booktitle={CVPR},
  pages={21666--21675},
  year={2024}
}

@inproceedings{RCM,
  title={Raising the Ceiling: Conflict-Free Local Feature Matching with Dynamic View Switching},
  author={Lu, Xiaoyong and Du, Songlin},
  booktitle={ECCV},
  pages={256--273},
  year={2024},
  organization={Springer}
}

@inproceedings{surf,
  title={Surf: Speeded up robust features},
  author={Bay, Herbert and Tuytelaars, Tinne and Van Gool, Luc},
  booktitle={ECCV},
  pages={404--417},
  year={2006},
  organization={Springer}
}

@article{schleiss2022vpair,
  title={VPAIR-Aerial Visual Place Recognition and Localization in Large-scale Outdoor Environments},
  author={Schleiss, Michael and Rouatbi, Fahmi and Cremers, Daniel},
  journal={arXiv:2205.11567},
  year={2022}}

@article{RK-Net,
  title={Joint Representation Learning and Keypoint Detection for Cross-view Geo-localization},
  author={Lin, Jinliang and Zheng, Zhedong and Zhong, Zhun and Luo, Zhiming and Li, Shaozi and Yang, Yi and Sebe, Nicu},
  journal={IEEE Transactions on Image Processing (TIP)},
  doi = {10.1109/TIP.2022.3175601},
  year={2022},
  }

@article{mughal2021assisting,
  title={Assisting UAV localization via deep contextual image matching},
  author={Mughal, Muhammad Hamza and Khokhar, Muhammad Jawad and Shahzad, Muhammad},
  journal={IEEE Journal of Selected Topics in Applied Earth Observations and Remote Sensing},
  volume={14},
  pages={2445--2457},
  year={2021},
  publisher={IEEE}
}

@article{FSRA,
  title={A transformer-based feature segmentation and region alignment method for UAV-view geo-localization},
  author={Dai, Ming and Hu, Jianhong and Zhuang, Jiedong and Zheng, Enhui},
  journal={IEEE TCSVT},
  volume={32},
  number={7},
  pages={4376--4389},
  year={2021},
  publisher={IEEE}
}

@article{DAC,
  title={Enhancing Cross-View Geo-Localization with Domain Alignment and Scene Consistency},
  author={Xia, Panwang and Wan, Yi and Zheng, Zhi and Zhang, Yongjun and Deng, Jiwei},
  journal={IEEE TCSVT},
  year={2024},
  publisher={IEEE}
}

@article{shen2023mccg,
  title={MCCG: A ConvNeXt-based multiple-classifier method for cross-view geo-localization},
  author={Shen, Tianrui and Wei, Yingmei and Kang, Lai and Wan, Shanshan and Yang, Yee-Hong},
  journal={IEEE TCSVT},
  year={2023},
  publisher={IEEE}
}

@inproceedings{s4g,
  title={Sample4geo: Hard negative sampling for cross-view geo-localisation},
  author={Deuser, Fabian and Habel, Konrad and Oswald, Norbert},
  booktitle={ICCV},
  pages={16847--16856},
  year={2023}
}

@article{wu2024camp,
  title={CAMP: A Cross-View Geo-Localization Method using Contrastive Attributes Mining and Position-aware Partitioning},
  author={Wu, Qiong and Wan, Yi and Zheng, Zhi and Zhang, Yongjun and Wang, Guangshuai and Zhao, Zhenyang},
  journal={IEEE Transactions on Geoscience and Remote Sensing},
  year={2024},
  publisher={IEEE}
}

@Article{liu2022convnet,
  author  = {Zhuang Liu and Hanzi Mao and Chao-Yuan Wu and Christoph Feichtenhofer and Trevor Darrell and Saining Xie},
  title   = {A ConvNet for the 2020s},
  journal = {CVPR},
  year    = {2022},
}

@article{P3P,
  title={Complete solution classification for the perspective-three-point problem},
  author={Gao, Xiao-Shan and Hou, Xiao-Rong and Tang, Jianliang and Cheng, Hang-Fei},
  journal={IEEE transactions on pattern analysis and machine intelligence},
  volume={25},
  number={8},
  pages={930--943},
  year={2003},
  publisher={IEEE}
}

@article{wang2018unmanned,
  title={Unmanned aerial vehicle oblique image registration using an ASIFT-based matching method},
  author={Wang, Chengyi and Chen, Jingbo and Chen, Jiansheng and Yue, Anzhi and He, Dongxu and Huang, Qingqing and Zhang, Yi},
  journal={Journal of Applied Remote Sensing},
  volume={12},
  number={2},
  pages={025002--025002},
  year={2018},
  publisher={Society of Photo-Optical Instrumentation Engineers}
}

@article{yuan2020automated,
  title={Automated accurate registration method between UAV image and Google satellite map},
  author={Yuan, Yijie and Huang, Wei and Wang, Xiangxin and Xu, Huaiyu and Zuo, Hongying and Su, Ruidan},
  journal={Multimedia Tools and Applications},
  volume={79},
  pages={16573--16591},
  year={2020},
  publisher={Springer}
}

@article{song2019oblique,
  title={Oblique aerial image matching based on iterative simulation and homography evaluation},
  author={Song, Woo-Hyuck and Jung, Hong-Gyu and Gwak, In-Youb and Lee, Seong-Whan},
  journal={Pattern Recognition},
  volume={87},
  pages={317--331},
  year={2019},
  publisher={Elsevier}
}

@inproceedings{DKM,
title={{DKM}: Dense Kernelized Feature Matching for Geometry Estimation},
author={Edstedt, Johan and Athanasiadis, Ioannis and Wadenbäck, Mårten and Felsberg, Michael},
booktitle={CVPR},
year={2023}
}

@article{roma,
title={{RoMa: Robust Dense Feature Matching}},
author={Edstedt, Johan and Sun, Qiyu and Bökman, Georg and Wadenbäck, Mårten and Felsberg, Michael},
journal={CVPR},
year={2024}
}

@inproceedings{omniglue,
  title={OmniGlue: Generalizable Feature Matching with Foundation Model Guidance},
  author={Jiang, Hanwen and Karpur, Arjun and Cao, Bingyi and Huang, Qixing and Araujo, Andr{\'e}},
  booktitle={CVPR},
  pages={19865--19875},
  year={2024}
}

@article{alike,
  title={Alike: Accurate and lightweight keypoint detection and descriptor extraction},
  author={Zhao, Xiaoming and Wu, Xingming and Miao, Jinyu and Chen, Weihai and Chen, Peter CY and Li, Zhengguo},
  journal={IEEE Transactions on Multimedia},
  volume={25},
  pages={3101--3112},
  year={2022},
  publisher={IEEE}
}

@inproceedings{potje2024xfeat,
  title={XFeat: Accelerated Features for Lightweight Image Matching},
  author={Potje, Guilherme and Cadar, Felipe and Araujo, Andr{\'e} and Martins, Renato and Nascimento, Erickson R},
  booktitle={CVPR},
  pages={2682--2691},
  year={2024}
}

@inproceedings{MI,
  title={Vision-based absolute localization for unmanned aerial vehicles},
  author={Yol, Aurelien and Delabarre, Bertrand and Dame, Amaury and Dartois, Jean-Emile and Marchand, Eric},
  booktitle={2014 IEEE/RSJ International Conference on Intelligent Robots and Systems},
  pages={3429--3434},
  year={2014},
  organization={IEEE}
}

@article{bianchi2021uav,
  title={UAV localization using autoencoded satellite images},
  author={Bianchi, Mollie and Barfoot, Timothy D},
  journal={IEEE Robotics and Automation Letters},
  volume={6},
  number={2},
  pages={1761--1768},
  year={2021},
  publisher={IEEE}
}

@article{yin2023isimloc,
  title={isimloc: Visual global localization for previously unseen environments with simulated images},
  author={Yin, Peng and Cisneros, Ivan and Zhao, Shiqi and Zhang, Ji and Choset, Howie and Scherer, Sebastian},
  journal={IEEE Transactions on Robotics},
  volume={39},
  number={3},
  pages={1893--1909},
  year={2023},
  publisher={IEEE}
}

@inproceedings{chen2021real,
  title={Real-time geo-localization using satellite imagery and topography for unmanned aerial vehicles},
  author={Chen, Shuxiao and Wu, Xiangyu and Mueller, Mark W and Sreenath, Koushil},
  booktitle={2021 IEEE/RSJ International Conference on Intelligent Robots and Systems (IROS)},
  pages={2275--2281},
  year={2021},
  organization={IEEE}
}

@article{chen2024fpi,
  title={OS-FPI: A Coarse-to-Fine One-Stream Network for UAV Geo-Localization},
  author={Chen, Jiahao and Zheng, Enhui and Dai, Ming and Chen, Yifu and Lu, Yusheng},
  journal={IEEE Journal of Selected Topics in Applied Earth Observations and Remote Sensing},
  year={2024},
  publisher={IEEE}
}

@inproceedings{sun2021loftr,
  title={LoFTR: Detector-Free Local Feature Matching with Transformers},
  author={Jiaming Sun and Zehong Shen and Yuang Wang and Hujun Bao and Xiaowei Zhou},
  booktitle = {CVPR},
  year={2021}
}

@article{LPN,
  title={Each part matters: Local patterns facilitate cross-view geo-localization},
  author={Wang, Tingyu and Zheng, Zhedong and Yan, Chenggang and Zhang, Jiyong and Sun, Yaoqi and Zheng, Bolun and Yang, Yi},
  journal={IEEE TCSVT},
  volume={32},
  number={2},
  pages={867--879},
  year={2021},
  publisher={IEEE}
}

@inproceedings{berton2022deep,
  title={Deep visual geo-localization benchmark},
  author={Berton, Gabriele and Mereu, Riccardo and Trivigno, Gabriele and Masone, Carlo and Csurka, Gabriela and Sattler, Torsten and Caputo, Barbara},
  booktitle={Proceedings of the IEEE/CVF Conference on Computer Vision and Pattern Recognition},
  pages={5396--5407},
  year={2022}
}

@article{chang2023review,
  title={A review of UAV autonomous navigation in GPS-denied environments},
  author={Chang, Yingxiu and Cheng, Yongqiang and Manzoor, Umar and Murray, John},
  journal={Robotics and Autonomous Systems},
  pages={104533},
  year={2023},
  publisher={Elsevier}
}

@article{chen2023oblique,
  title={An oblique-robust absolute visual localization method for GPS-denied UAV with satellite imagery},
  author={Chen, Yuan and Jiang, Jie},
  journal={IEEE Transactions on Geoscience and Remote Sensing},
  year={2023},
  publisher={IEEE}
}

@ARTICLE{ORB-SLAM3,
  author={Campos, Carlos and Elvira, Richard and Rodríguez, Juan J. Gómez and M. Montiel, José M. and D. Tardós, Juan},
  journal={IEEE Transactions on Robotics}, 
  title={ORB-SLAM3: An Accurate Open-Source Library for Visual, Visual–Inertial, and Multimap SLAM}, 
  year={2021},
  volume={37},
  number={6},
  pages={1874-1890},
  doi={10.1109/TRO.2021.3075644}}

@misc{interim_regulations_uav,
  title        = {Interim Regulations on the Flight Management of Unmanned Aerial Vehicles},
  howpublished = {\url{https://www.gov.cn/zhengce/content/202306/content_6888799.htm}},
  year={2023},
organization={THE STATE COUNCILTHE PEOPLE'S REPUBLIC OF CHINA}
}

@inproceedings{gurgu2022vision,
  title={Vision-based gnss-free localization for uavs in the wild},
  author={Gurgu, Marius-Mihail and Queralta, Jorge Pe{\~n}a and Westerlund, Tomi},
  booktitle={2022 7th International Conference on Mechanical Engineering and Robotics Research (ICMERR)},
  pages={7--12},
  year={2022},
  organization={IEEE}
}

@inproceedings{dedode,
  title={DeDoDe: Detect, Don’t Describe—Describe, Don’t Detect for Local Feature Matching},
  author={Edstedt, Johan and B{\"o}kman, Georg and Wadenb{\"a}ck, M{\aa}rten and Felsberg, Michael},
  booktitle={2024 International Conference on 3D Vision (3DV)},
  pages={148--157},
  year={2024},
  organization={IEEE}
}

@article{GE_simulator,
  title={Sequence matching for Image-Based UAV-to-Satellite Geolocalization},
  author={Wang, Zhen and Shi, Dianxi and Qiu, Chunping and Jin, Songchang and Li, Tongyue and Shi, Yanyan and Liu, Zhe and Qiao, Ziteng},
  journal={IEEE Transactions on Geoscience and Remote Sensing},
  year={2024},
  publisher={IEEE}
}

@article{he2023foundloc,
  title={Foundloc: Vision-based onboard aerial localization in the wild},
  author={He, Yao and Cisneros, Ivan and Keetha, Nikhil and Patrikar, Jay and Ye, Zelin and Higgins, Ian and Hu, Yaoyu and Kapoor, Parv and Scherer, Sebastian},
  journal={arXiv preprint arXiv:2310.16299},
  year={2023}
}

@article{he2024leveraging,
  title={Leveraging Map Retrieval and Alignment for Robust UAV Visual Geo-Localization},
  author={He, Mengfan and Liu, Jiacheng and Gu, Pengfei and Meng, Ziyang},
  journal={IEEE Transactions on Instrumentation and Measurement},
  year={2024},
  publisher={IEEE}
}

@article{ye2024coarse,
  title={A coarse-to-fine visual geo-localization method for GNSS-denied UAV with oblique-view imagery},
  author={Ye, Qin and Luo, Junqi and Lin, Yi},
  journal={ISPRS Journal of Photogrammetry and Remote Sensing},
  volume={212},
  pages={306--322},
  year={2024},
  publisher={Elsevier}
}

@inproceedings{University-1652,
  title={University-1652: A multi-view multi-source benchmark for drone-based geo-localization},
  author={Zheng, Zhedong and Wei, Yunchao and Yang, Yi},
  booktitle={Proceedings of the 28th ACM international conference on Multimedia},
  pages={1395--1403},
  year={2020}
}

@article{sues200,
  title={SUES-200: A multi-height multi-scene cross-view image benchmark across drone and satellite},
  author={Zhu, Runzhe and Yin, Ling and Yang, Mingze and Wu, Fei and Yang, Yuncheng and Hu, Wenbo},
  journal={IEEE TCSVT},
  volume={33},
  number={9},
  pages={4825--4839},
  year={2023},
  publisher={IEEE}
}

@InProceedings{D2-Net,
    author = {Dusmanu, Mihai and Rocco, Ignacio and Pajdla, Tomas and Pollefeys, Marc and Sivic, Josef and Torii, Akihiko and Sattler, Torsten},
    title = {{D2-Net: A Trainable CNN for Joint Detection and Description of Local Features}},
    booktitle = {CVPR},
    year = {2019},
}

@article{DenseUAV,
  title={Vision-based UAV self-positioning in low-altitude urban environments},
  author={Dai, Ming and Zheng, Enhui and Feng, Zhenhua and Qi, Lei and Zhuang, Jiedong and Yang, Wankou},
  journal={IEEE Transactions on Image Processing},
  year={2023},
  publisher={IEEE}
}

@article{UAV-VisLoc,
  title={UAV-VisLoc: A Large-scale Dataset for UAV Visual Localization},
  author={Xu, Wenjia and Yao, Yaxuan and Cao, Jiaqi and Wei, Zhiwei and Liu, Chunbo and Wang, Jiuniu and Peng, Mugen},
  journal={arXiv preprint arXiv:2405.11936},
  year={2024}
}

@article{couturier2021review,
  title={A review on absolute visual localization for UAV},
  author={Couturier, Andy and Akhloufi, Moulay A},
  journal={Robotics and Autonomous Systems},
  volume={135},
  pages={103666},
  year={2021},
  publisher={Elsevier}
}

@inproceedings{ji2024game4loc,
    title={Game4loc: A uav geo-localization benchmark from game data},
  author={Ji, Yuxiang and He, Boyong and Tan, Zhuoyue and Wu, Liaoni},
  booktitle={Proceedings of the AAAI Conference on Artificial Intelligence},
  volume={39},
  number={4},
  pages={3913--3921},
  year={2025}
}

@inproceedings{Wu2023uav4l,
    title={UAVD4L: A Large-Scale Dataset for UAV 6-DoF Localization}, 
    author={Rouwan Wu and Xiaoya Cheng and Juelin Zhu and Xuxiang Liu and Maojun Zhang and Shen Yan},
    booktitle={International Conference on 3D Vision (3DV)},
    year={2024}
}
}


\end{document}